\documentclass[runningheads]{llncs}
\usepackage[utf8]{inputenc} 
\usepackage{graphicx}
\usepackage{amsmath,amssymb} % define this before the line numbering.
\usepackage{color}
\usepackage[width=122mm,left=12mm,paperwidth=146mm,height=193mm,top=12mm,paperheight=217mm]{geometry}
\usepackage{subfig}
\usepackage{hyperref}
\usepackage{float}

\begin{document}
% \linenumbers
\pagestyle{headings}
\mainmatter
\def\ECCV18SubNumber{2890}  % Insert your submission number here

\title{Person re-identification across different datasets with multi-task learning} % Replace with your title

%\titlerunning{ECCV-18 submission ID \ECCV18SubNumber}

%\authorrunning{ECCV-18 submission ID \ECCV18SubNumber}

%\author{Anonymous ECCV submission}
%\institute{Paper ID \ECCV18SubNumber}
\author{Matthieu Ospici, Antoine Cecchi}
\institute{Atos BDS R\&D \\
  \email{\{matthieu.ospici,antoine.cecchi\}@atos.net}} 

\maketitle

\begin{abstract}
  This paper presents an approach to tackle the re-identification problem.  This  is a challenging problem due to the large variation of pose, illumination or camera view.
  More and more datasets are available to train machine learning models for person re-identification. These datasets vary in conditions:  cameras numbers, camera positions, location, season, in size, i.e. number of images, number of different identities. Finally in labeling: there are datasets annotated with attributes while others are not.
  To deal with this variety of datasets we present in this paper an approach to take information from different datasets to build a system which performs well on all of them. Our model is based on a Convolutional Neural Network (CNN) and trained using multitask learning. Several losses are used to extract the different information available in the different datasets. Our main task is learned with a classification loss.  To reduce the intra-class variation we experiment with the center loss \cite{center-loss}.
  Our paper ends with a performance evaluation in which we discuss the influence of the different losses on the global re-identification performance.  We show that with our method, we are able to build a system that performs well on different datasets and simultaneously extracts attributes. We also show that our system outperforms recent re-identification works on two datasets.

\end{abstract}

\section{Introduction}

\begin{figure}
\centering
  \subfloat[VIPeR]{\includegraphics[width=3cm,height=4cm]{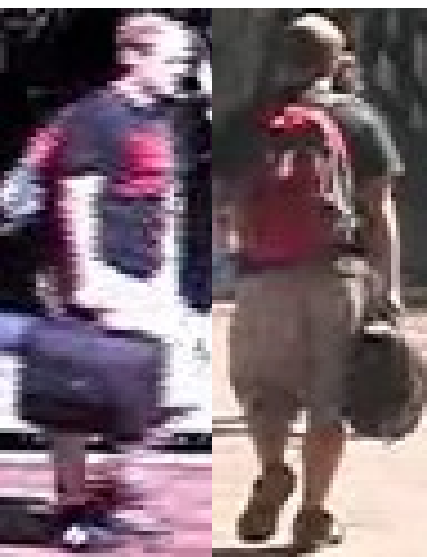}\label{fig:viper_ex}}
  \hspace{0.7cm}
  \subfloat[CUHK01]{\includegraphics[width=3cm,height=4cm]{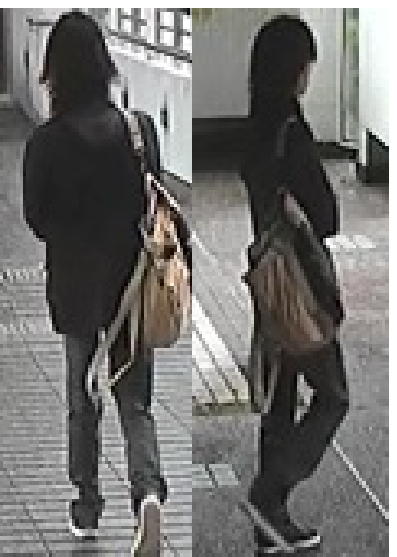}\label{fig:cuhk01_ex}}
  \hspace{0.7cm}
  \subfloat[CUHK03]{\includegraphics[width=3cm,height=4cm]{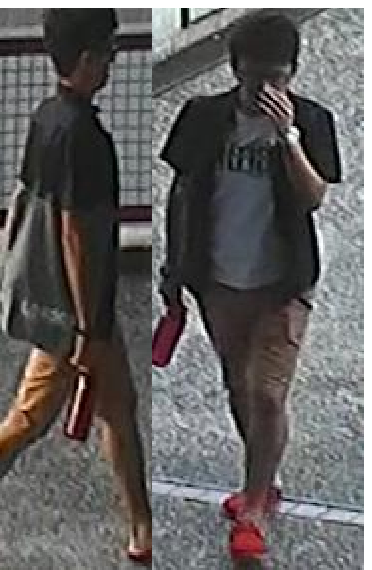}\label{fig:cuhk03_ex}}
  %\vspace{0.5cm}
   \caption{Three re-identification datasets used in our work.}\label{fig:ex_ds}
\end{figure}

In many domains, such as surveillance or digital signage, being able to automatically recognize a person across different, non-overlapping cameras, without the help of a human operator is very valuable. This task is known as person re-identification and can be extremely challenging since great variations can occur between the different cameras. Figure~\ref{fig:ex_ds} shows two images, taken from two different cameras from three academic datasets: VIPeR \cite{viper-ds}, CUHK01 \cite{cuhk01} and CUHK03 \cite{chhk03}.  Variation can be large between two pictures belonging to the same dataset such as body pose, luminosity, view angle or background.

In many works, person re-identification is based on a similarity score between a pair of images. If the two images represent the same person, the similarity score is high. Two aspects are usually studied. The first one consists in extracting robust invariant features to represent the appearance of a person \cite{liu2012,Madden2007,farenzena2010}. The second is metric learning \cite{dick2010,LiaoHZL15}: it consists in learning the best possible metric to discriminate between positive and negative samples.

Recently, convolutional neural networks demonstrated very high efficiency in several computer vision problems such as image segmentation \cite{fcn} or object recognition \cite{NIPS2012_alex}. Many research projects have proved that deep neural networks  are also extremely efficient for re-identification \cite{combination,market1501,zheng2016person,attribute-comple}.

To train such a deep neural network, large datasets are mandatory. Recently, re-identification datasets large enough to train deep models have emerged \cite{market1501,chhk03,mars-ds}. In many works \cite{eccv-train-not-join,attribute-comple}, a neural network is trained on a large dataset and then fine tuned on a smaller one. Consequently, for the performance evaluation, a specific fine tuned model is used to evaluate its corresponding dataset. For an industrial purpose, having a single model able to perform well on many datasets is extremely important. It means that the model can handle different situations, which enables deploying the same model on cameras installed in different environments.

Re-identification with CNNs is usually performed using features extracted by the neural network from identities during the training phase. Attributes, that are more high-level features, like gender, clothes length, handbag may be extremely valuable for re-identification since such features are truly robust to view-angle and cameras change. \emph{Schumann et al.} \cite{attribute-comple} demonstrated that using only attributes leads to low performance compared to the features learned by a CNN from the identities. A good approach is therefore to use a combination of attributes and features extracted from identities.

To have a system able to make use of attributes, an access to a large dataset annotated with attributes is required. Nevertheless, it is difficult to acquire large training data for a set of attributes since manual annotations is extremely expensive. Thus, only a subset of re-identification datasets is annotated with attributes and many of them will remain attributeless. It is therefore a problem to build a general system that performs well on several datasets and make use of attributes. To deal with the variation of size and annotation of re-identification datasets,  we present in this paper a multi-task learning approach which learns the re-identification task from a combination of several datasets. Furthermore, our system is able to take advantage of attribute information in dataset annotated with.

Two main strategies are used in the re-identification community for training deep neural networks. We will describe them in more detail in the next section. The first one \cite{siamese-reid,duff,quadruplet} is based on siamese networks, contractive or triplet losses. The second one \cite{guided-dropout,zheng2016person}, used in our work, is based on classification losses. Since the last layer is a linear classifier, classification methods ensure that features are linearly separable. Consequently, the distance between features belonging to two different classes increases. Nevertheless, with this approach, the intra-class variation is not controlled. Intuitively, reducing the intra-class variations can make the features more discriminant and then increase the re-identification performance. In this work, we add one task in our multi-task learning objective: a task designed to force the features of same identities to be the closest as possible. For the implementation, we employ a method described in \cite{center-loss} which proposes a loss called \emph{center loss}. We then evaluate the interest of this center loss for our re-identification system.

The contributions of our work are three folds:

\begin{itemize}
\item We build a model that learns a generic representation of the person using several datasets for re-identification (CHUK01 \cite{cuhk01}, CHUK03 \cite{chhk03}, MARS \cite{mars-ds}, ViPER \cite{viper-ds}, Market1501 with attributes \cite{market150-att}) to build a system that performs well on all of these datasets without performing a specific fine tuning.
\item We take advantage of attributes available in some re-identification datasets such as hair length, top/bottom color, clothes length. We have a multi-task learning objective: the re-identification task, learned from all the datasets and the attribute classification tasks, learned from a subset of the available datasets.
\item We evaluate an auxiliary task designed to control the intraclass variation of the re-identification features. This task is based on the center loss described in \cite{center-loss}.

\end{itemize}

%1.  from several dataset which perform well on all of them.
%2. We use a center loss [] to decrease the intraclass variation which allow a training procedure simpler than triplet loss.
%3. The same network is able to output both a generic representation and a list of attributes for each human. We evaluate the quality of the attribute and the ability of the attributes to increase the quality of the generic representation.

\section{Related work}

\textbf{Person re-identification} Many studies are lead on  re-identification, and today CNN and deep learning approaches are well studied and show very good performance on many datasets \cite{reid-cvpr15,guided-dropout,duff,ZhengZY18}.

Two types of approaches are usually chosen for recent re-identification works with CNN.  First one is based on siamese networks \cite{sia-ref,siamese-reid}, triplet loss networks \cite{facenet,duff}, quadruplet loss \cite{quadruplet} to learn a representation based on different and identical couple/triplet. The other approach is based on identity losses \cite{guided-dropout,mars-ds}, in which each identity is seen as a class. A classification loss function, such as softmax cross entropy is usually employed. 

Training is different whether one chooses the first or the second approach. Triplet loss networks can be difficult to train since one needs to preprocess data to find triplets and hard positive and negative samples \cite{facenet}. Compared to the number of samples in the dataset, the couple/triplet needs for training dramatically increase and can lead to slow convergence.
%With the classification approach, 

% In similar problem a center loss\cite{center-loss} which learn a center for each classes during training and constraint each sample to be close to a center. The effect is to reduce the intra-class variations.

\emph{Ahmed et al.} \cite{reid-cvpr15} takes two images for both train and test time to be able to decide whether the two images represent the same person or not. This approach needs to perform an inference each time we need to compare two images, which requires lot of computing power during a search.

%A large part of recent reidentification works [cite,cite] extract a representation using deep learning and CNN. A metric between these representation is used to perform the re-identification task. %Our work has a similar approach, we extract from an image a vector of 4096 dimensions. Cosine distance between the vector should be small if two images represent the same person, and cosine distance should be large if the two persons are different.

To deal with several datasets \emph{Xiao et al.} \cite{guided-dropout} developed a guided dropout strategy  to learn person representation across different dataset. Other approaches  specialize a trained network to a particular dataset. For example, because a large CNN needs many samples, many works fine-tune a network trained on a large re-identification dataset to a smaller one to not overfit \cite{eccv-train-not-join,attribute-comple,reid-cvpr15}.

\textbf{Multi-task learning for person re-identification}

Multi-task learning \cite{MTL} has been applied to re-identification. For example, in \cite{mtl-reid} the authors use a siamese network, the different tasks are attributes classification tasks. In \cite{mtl-reid-cam}, re-identifications from multiple cameras are regarded as related tasks.
Some approaches use a network with two branches \cite{deep-joint-2017,attribute-comple}, with jointly optimized losses.

\textbf{Attributes for person re-identification}  

Attributes have been extensively studied in re-identification \cite{layne-reid-att,layne-reid-att2,Layne2014}. Attributes can preserve robust information of a person across different point of views or conditions and then it is natural to use them for re-identification. More recently, attributes have been used with deep learning approaches. These architectures can be trained with relatively large re-identification datasets annotated with attributes \cite{peta-ds,market1501}. Some researchers use architectures with two branches, one for the re-identification  and one for the attribute extraction and combine them \cite{duff,combination}. \emph{Su et al} \cite{Sueccv16} use three stages of fine tuning and a triplet based loss. \emph{Matsukawa et. al} \cite{combination} only use a combination of losses based on attributes to create a representation able to perform re-identification.

%In our approach, we train a CNN  with a classification loss, contrary to the work using classification losses, we control the intra-classes variation using the center loss. % To our knowledge, no other published work experiment the center loss on the person re-identification problem. %Our multi task learning approach is close to \cite{mtl-reid-cam}, we have the main classification task and $N$ auxiliary tasks based on attributes. Nevertheless, our approach is able to deal with several kind of datasets, some with attributes and the other without.

%TODO il faut etoffer cette partie

%We train our model with a combination of several dataset. We take care of the classes imbalances between the different dataset with weighted  cross entropy. 

\section{Proposed approach}

In this paper we present a deep learning approach for the problem of person re-identification. Given an image with a person, the network outputs a global representation of this person. This representation should be independent from the person pose. We call it the signature of the person. Furthermore, in our architecture, the same network also outputs a list of attributes. The complete attributes list we support is detailed in Table~\ref{tab:att}. To decide if two pictures represent the same person, we compute the cosine distance between the two signatures. The smallest the distance is, the more likely it is for the two signatures to represent the same person.

\begin{table}
  \centering
  \caption{Attributes supported by our system}
\begin{tabular}{ l l }
\hline
Attribute & Possible Values \\
\hline
genre & (male, female) \\
top color & (black ,blue, green, grey, purple, red, white, yellow) \\
bottom color & (black ,blue, brown, grey, green, pink, purple, white, yellow) \\
top length & (long, short) \\
bottom length & (long, short) \\
backpack & (true, false) \\
hand bag & (true, false) \\
other bag & (true, false) \\
hair length & (long, short) \\
\hline
\end{tabular}
\label{tab:att}
\end{table}

\subsection{Model design}
\begin{figure}
\centering
\includegraphics[width=12cm,height=6cm]{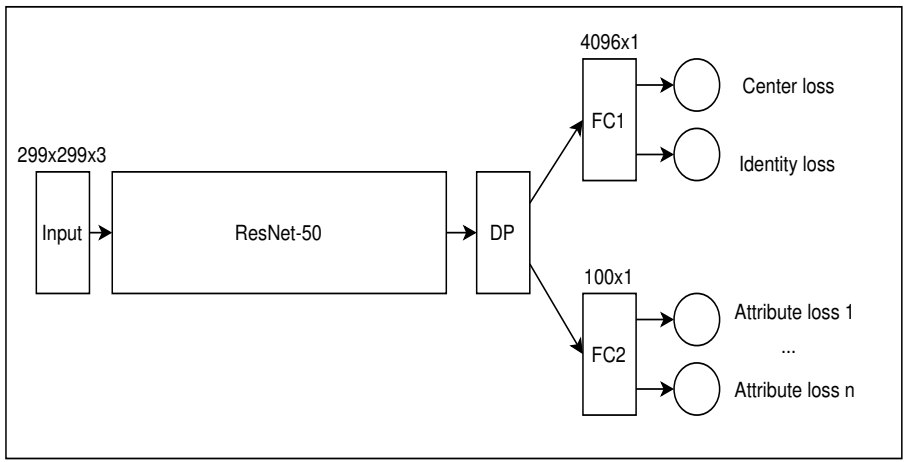}
\caption{The CNN architecture for the re-identification. A resnet50 \cite{resnet50} network is followed by two fully-connected layers. The FC1 layer is used to compute the similarity distance and it is trained with the center loss and an identity loss. FC2 is trained with the attribute losses. A dropout layer (DP) is added before FC1 and FC2.}
\label{fig:net-base}
\end{figure}

The network architecture used in our work is represented Figure~\ref{fig:net-base}. It is based on the resnet50 \cite{resnet50} model followed by a dropout (DP) and two fully connected layers, $FC1$ and $FC2$. 

\begin{description}
\item[FC1] This layer is the global person representation (the signature). Therefore, it is used to perform the re-identification. In our architecture, its size is set to $4096$.
  Two losses are used to train this layer. One identity loss and the center loss \cite{center-loss}.

  The objective of the identity loss is to make the network able to classify each identity into the correct class. It is a multi-class classification,  therefore the cross-entropy softmax loss is used.

  %\subsubsection{Controlling the intraclass variation with the center loss}
%Pourquoi seulement 2 losses ? 
  The identity loss  forces the deep features to be separable. To reduce the intra class variation, we use the center loss, introduced by \emph{Wen et al.} \cite{center-loss} that organizes the deep features around a center for each classes. During the training, the centers are learned and the distance between the deep features and their corresponding center are minimized.  \emph{Wen et al} shown that the center loss is differentiable: our deep neural network can therefore  be trained with a standard algorithm based on gradient descent.
As stated in \cite{center-loss}, the center loss is given by (\ref{cs_eq}), in which $m$ is the size of the batch,  $c_{y_i}$ is the center of the $y_i$ class and $x_i$ is deep feature. Dimensions of both $x_i$ and $c_{y_i}$ are the dimension of FC1: $4096$.
  
  \begin{equation}
 L_{cs}=\sum_{i=1}^{m} ||x_i-c_{y_i}||_2^2
 \label{cs_eq}
\end{equation}

  The balance between the identity loss and the center loss is done by a $\alpha$ factor.

\item[FC2] Our system is trained with attributes. We support several attributes and, for each of them, we train a classifier. Two types of classifiers are used for the two types of attributes. For the binary attribute (e.g. male/female) binary classifiers are used and trained with a sigmoid cross entropy. For the multi-class attributes (e.g. top/bottom colors) a softmax cross entropy loss is used.
To avoid corrupting the global representation of FC1, we choose to connect the attributes classifiers on FC2, which is a $100$ dimensions layer. The weight of the attributes losses is controlled by a $\lambda$ parameter.

\end{description}

\subsubsection{Multi-task learning}

The learning objectives are controlled by several losses, two for FC1 and $n$ for FC2, with $n$ the number of attributes. As showed in Figure~\ref{fig:multi}, the learning is performed by the combination of all these losses.

Let's $L_{id}$ the identity loss, $L_{cs}$ the center loss and $L_{att}$ the sum of the attribute losses, our total loss is given by (\ref{loss_total})

  \begin{equation}
 L_{total}=L_{id} + \alpha.L_{cs} + \lambda.L_{att}
 \label{loss_total}
\end{equation}

 \subsubsection{Re-identification process}

The re-identification process is based on the global representation FC1 extracted from pictures. The similarity between pictures is computed with the cosine distance.

\begin{equation}
   d_{cos}(a,b) = 1-\dfrac{<a,b>}{||a||_2.||b||_2}
   \label{cosine distance}
\end{equation}
In which $a$ and $b$ are two $4096$ dimension vectors extracted from two pictures by the system. And $<.,.>$ represent the euclidean dot product.

%Final layers, output a signature, a vector of $4096$ elements which is the generic representation of the person given in input.

\newpage

\begin{table}
  \centering
  \caption{Datasets used on our work.}
\begin{tabular}{ l c c c c }
\hline
Dataset & classes & Training & Test & Attributes \\
\hline
CUHK01 \cite{cuhk01} & 971 & 1552 & 388 & No\\
CUHK03 \cite{chhk03} & 1467 & 21012 & 5252 & No\\
VIPeR  \cite{viper-ds}& 632 & 506 & 126 & No\\
MARKET 1501 \cite{market1501}& 1501 & 12936 & 19732 & \emph{Yes} \\
\hline
\end{tabular}
\label{tab:ds}

\end{table}

\subsection{Learning strategy}
\label{LS}

The aim of our work is to build a single system able to perform well simultaneously on all the datasets listed on Table~\ref{tab:ds} and to recognize pedestrian attributes. We describe in this section the way we chose to reach this goal. 

\subsubsection{Joint datasets learning} 
\label{jdl}
One of the main objectives of our system is to show good performances on diverse datasets. Thus to build our training set, we proceed by mixing the training sets of these datasets. %Un peu répétitif
This approach is valid since there is no identity overlap between the different datasets. Let's consider we have $n$ datasets, the number of identity of the $i\textsuperscript{th}$ is denoted by $d_i$. The total number of identities of our global dataset is $\sum\limits_{i=0}^{n-1} d_i$.

The identities in the datasets are not represented by the same number of images. Thus one can not simply merge the datasets since the information contained in the smallest datasets would then be negligible compared to that contained in the large ones. To tackle this issue we employe a weighted cross entropy for the identity loss.

Let $nc_i^k$ represent the number of images in the $k_{th}$ class of the $i_{th}$ dataset. $q_i^k$ represents the value of the corresponding logit and $p_i^k$ the associated ground truth. The identity loss $l_{id}$ is thus given by
\begin{equation}
 l_{id}=\sum_{i=0}^{n-1} \sum_{k=0}^{nc_i^k-1} \frac{1}{nc_i^k}p_i^k.log(q_i^k)
 \label{weighted_cross_entropy}
\end{equation}

This ensure a high weight for the classes under represented and an lower weigh for the most frequent ones. This loss is also appropriate for an optimization in batch mode, in which we compute the weighted loss for each element of the batch and we compute the mean over all the cross entropies. Let  $N_b$ be the size of the batch we denote the loss corresponding to the $g_{th}$ element of the batch  by $l_{id}^{(g)}$ defined as in~\eqref{weighted_cross_entropy}. One can then write the final loss $L_{id}$: 
\begin{equation}
 L_{id}=\frac{1}{N_b}\sum_{g=0}^{N_b-1} l_{id}^{(g)}
 \label{batch_loss}
\end{equation}

The training is done with the Adam optimizer \cite{adam}. The initial learning rate is set to $0.0001$. We start the training of our system with a resnet50 \cite{resnet50} network pretrained on imagenet.

%We also show on the evaluation performance part how performs the joint learning strategy compared to a single learning strategy in which we train a neural network only on one dataset and we test it on the test dataset.

\subsubsection{Attribute recogntion task}

To be able to have only one network able to output both re-identification and attributes, multi-task learning is used. We use three losses: the ones defined in~\eqref{weighted_cross_entropy} and in~\eqref{cs_eq} for the re-identification task and another for the attributes extraction task. As for the identity loss, the attribute loss can be written as a modified cross entropy. Let suppose there are $K$ annotated classes in our dataset, with $N_{att}$ attributes. The loss corresponding to the 
$l_{th}$ attribute of the $j_{th}$ class is written $l_{a}^{(j,l)}$ and is: 
\begin{equation}
 l_{a}^{(j,l)} = \sum_{d=0}^{d_{att}^l-1} p_{d}.log(q_{d})
 \label{one_att_loss}
\end{equation}
Where $d_{att}^l$ represents the dimension of the logit layer of the $l_{th}$ attribute, $p_d$ the output of the $d_{th}$ logit and $q_d$ the corresponding ground truth. One can then write the loss relative to the attributes: 
\begin{equation}
l_{att} = \sum_{j=0}^{K-1} \sum_{l=0}^{N_{att}-1} l_{a}^{(j,l)}
 \label{att_loss}
\end{equation}

\begin{figure}[ht]
\centering
\fbox{\includegraphics[width=12cm]{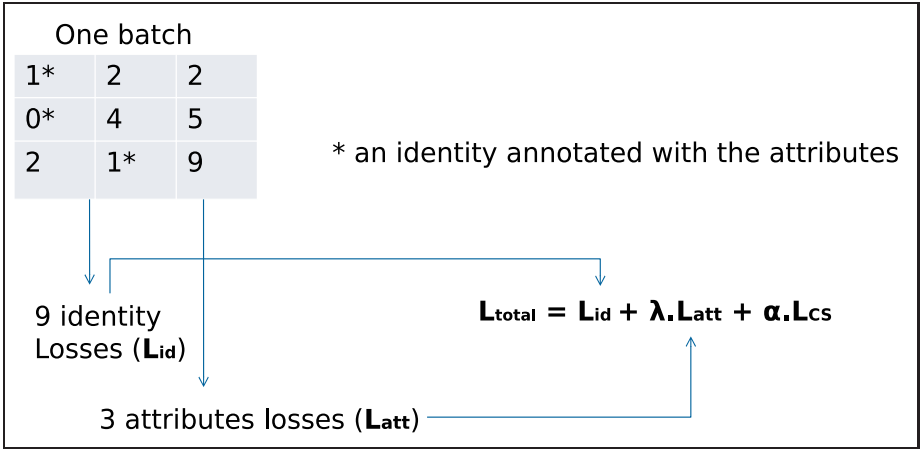}}
\caption{Multi-task learning implementation during training for a batch. $L_{id}$ is the identity loss, $L_{cs}$ the center loss and $L_{att}$ the sum of the attribute losses.  }
\label{fig:multi}
\end{figure}

During training, we randomly sample batches of images. While all the images are annotated for the re-identification task, only some of them are annotated with attributes. Therefore the re-identification loss is computed with all the batch samples, the attributes loss is updated on a subset of the batch. More details can be found on Figure~\ref{fig:multi}, in which each sample of a batch has an identity used for the re-identification task. 
Some of them also have attributes annotations. The identity loss is thus always computed on all the samples of the batch. On this example, the identity loss is computed on the 9 batch samples. The attribute loss is only computed with the samples having attributes annotation. In the batch given in example, there are only 3 samples annotated.
%In our implementation, the $\lambda$ factor controls the balance between the attributes loss and the re-identification loss.  $\lambda$ is sat to $100$ in our model. 

%\newpage

Lets define 
\begin{equation}
 A =  \{\beta_i , i \in \{0,1,...,N_b-1\}\} \in \{0,1\}^{N_b}
 \label{has_att}
\end{equation}

With $N_b$ representing the batch size as in~\eqref{batch_loss}. The loss $L$ defined in the figure \ref{fig:multi} can the be written as follows: 
\begin{equation}
 L = L_{id} + \alpha.L_{cs} + \lambda.\sum_{i=0}^{N_b-1} \beta_i.l_{att}^{(i)} 
 \label{loss_id_att}
\end{equation}

However in the dataset annotated with attributes, appearance frequencies of each attribute are not equal. For example, the blue pants class is more represented than the pink pants in the dataset. To deal with this unbalanced dataset a penalty in introduced in the loss. Such as for~\ref{jdl} the loss for a specific attribute~\eqref{one_att_loss} is weighted to penalize the most represented classes. Let $N_{ac}^{(d,l)}$ represent the number of occurrences of the $d_{th}$ class of the $l_{th}$ attribute. One can then re-write~\eqref{one_att_loss}:
\begin{equation}
 l_{a}^{(j,l)} = \sum_{d=0}^{d_{att}^l-1} \frac{1}{N_{ac}^{(d,l)}}.p_{d}.log(q_{d})
 \label{weighted_att}
\end{equation}

\section{Performance evaluation}

Our model is implemented using the TensorFlow library \cite{tensorflow2015}.

We perform our experiments on four re-identification datasets publicly available. The datasets used are  CHUK01 \cite{cuhk01}, CHUK03 \cite{chhk03},  ViPER \cite{viper-ds}, Market1501 with attributes \cite{market150-att}.

%To evaluate our performance we compute the Cumulative Match Characteristic (CMC) at rank $n$. CMC at rank $n$ is the probability that the correct match occurs with $n$ pictures of the gallery.
%Probe image is taken from one camera and the gallery set is taken from another camera.

We first present these datasets and the protocol followed to compute the performance metrics. Then we show the results of our approach and compare them to the state of the art.

\subsection{Datasets}

\begin{description}
\item[CUHK01] Dataset with 3,884 images of 972 pedestrians, each identity is observed by two cameras view. Each person has two images from the first camera and two images from the second camera. 
All pedestrian images are manually cropped
\item[CUHK03] dataset  contains  14,097  cropped  images  of 1,467 identities. 
Each identity is observed by two camera views and contains 4.8 images in average for  each  view. 
There are two  types  of  bounding boxes: the manually labeled pedestrian bounding boxes and the automaticaly detected bounding boxes obtained by a  pedestrian detector.% Both the single-shot and multiple-shot results will be reported.
\item[VIPeR] dataset is a very challenging dataset, since it contains 632 pedestrian image pairs taken from arbitrary viewpoints with various illumination conditions or poses. 
\item[Market 1501] This dataset contains 32668 images annotated using a DPM (Deformable Part Model) giving 1501 identities, split in a training set of 751 identities and a test set of 750 identities. Each identity is captured 
by at most 6 and at least 2 so that cross view search can be performed. Furthermore one can focus on the search from one viewpoint in each other. The dataset has been annotated with 27 attributes per ID. 
\end{description}

The statistics of this four datasets are summarized on Table~\ref{tab:ds}.

\subsection{Metric and protocol}

\subsubsection{Metric}
To evaluate our model we employed the Cumulative Matching Characteristic (CMC) \cite{Gray07evaluatingappearance} which is the most used metric on re-identification works \cite{mars-ds,viper-ds,cuhk01,duff,mtl-reid}. The  CMC  curve  
represents the probability of correct re-identification on the y-axis against the number of candidates  returned  on the x-axis. The CMC rank1 is very important since it measures the ability of the  system to truly
identify a person.

For the different datasets we use the protocol described in \cite{guided-dropout} which is based on \cite{Paisitkriangkrai2015LearningTR}. For CUHK01 and VIPeR we divide the identities of the dataset in two equal parts, i.e 485 and 316,
for the test set and the training set. For CUHK03, we use the commonly used split: 1467 identities for the training set and 100 identities for the test set.

Our gallery sets and probe sets are constructed as follows. For VIPeR, which has two camera views, we randomly select an image from the first camera as probe image. The gallery image is the same identity taken from the other camera. For 
CUHK03 and CUHK01 we use a similar protocol, we ensure that images of the gallery sets and probe sets are not the same camera. As stated in \cite{guided-dropout,Paisitkriangkrai2015LearningTR} both the manually and automatically 
cropped images were used in our experimentation. %We mainly refer in \cite{guided-dropout} for our experimentation since the authors built a system able to perform well on a combination of datasets.
\subsubsection{Training protocol}

We train several networks to compare the influence of the center loss and the attributes loss.

For the first stage, we train networks without the attributes losses. The resnet50 network we use is pretrained on imagenet. The weights are the one distributed by the Tensorflow community. 
The name of the checkpoint is \emph{resnet\_ v2\_50\_2017\_04\_14.tar.gz} and can be downloaded from the Tensorflow GitHub repository\footnote{\url{https://github.com/tensorflow}}. 
The learning is performed using the Adam \cite{adam} optimizer set with an initial learning rate at $10^{-4}$ for this stage. We train 4 networks, each network is trained with a particular center loss value ($0.0$, $0.05$, $0.06$ and $0.1$).

For the second stage we load the weight previously computed and we launch a train  with the Adam optimizer with learning rate sat to $10^{-6}$. We generate two types of models, the first type are models still trained without the 
attributes losses and the second type are models trained with the attributes losses.
At the end of these two stages we therefore have generated 10 models, for the 5 center loss values we have a model with attributes and a model without attributes.

For the training, the hyperparameters we used are available int Table~\ref{tab:hyp}.

\begin{table}[ht]
  \centering
\caption{Hyperparameters used in our model. $CS_{\alpha}$ is the learning rate for the center loss as described in \cite{center-loss}.}
\begin{tabular}{ l c }
\hline
Hyperparameter & Value \\
\hline
Dropout  & 0.8 \\
L2 regularization & 0.001 \\
Batch size & 64 \\
$\lambda$ & 100 \\
$CS_{\alpha}$ & 0.9 \\
$\alpha$ & $0.0$, $0.05$, $0.06$ and $0.1$ \\
\hline
\end{tabular}

\label{tab:hyp}
\end{table}

We discuss in this section both the influence of the center loss and the attributes. We first discuss only the center loss. To show its influence, the models are trained without attributes. Then we show the influence of the attributes. We therefore use our models trained with the attributes and the center loss.

\subsection{Influence of the center loss}

On this section we vary the $\alpha$ parameter which controls the center loss weight on the global loss.

\begin{figure}[ht]
\centering
  \subfloat[CMC train]{\includegraphics[width=5cm]{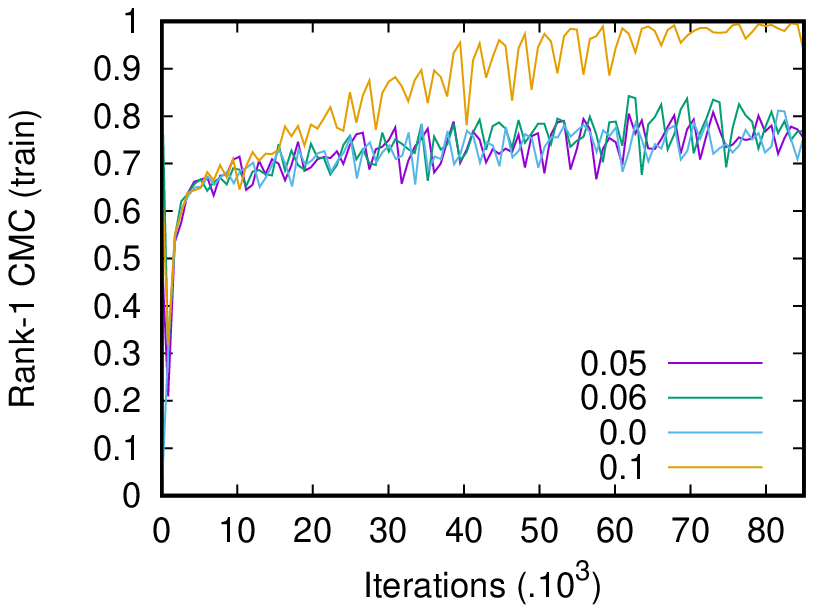}\label{fig:cmc_train}}
  \hspace{0.7cm}
  \subfloat[Center loss value during train]{\includegraphics[width=5cm]{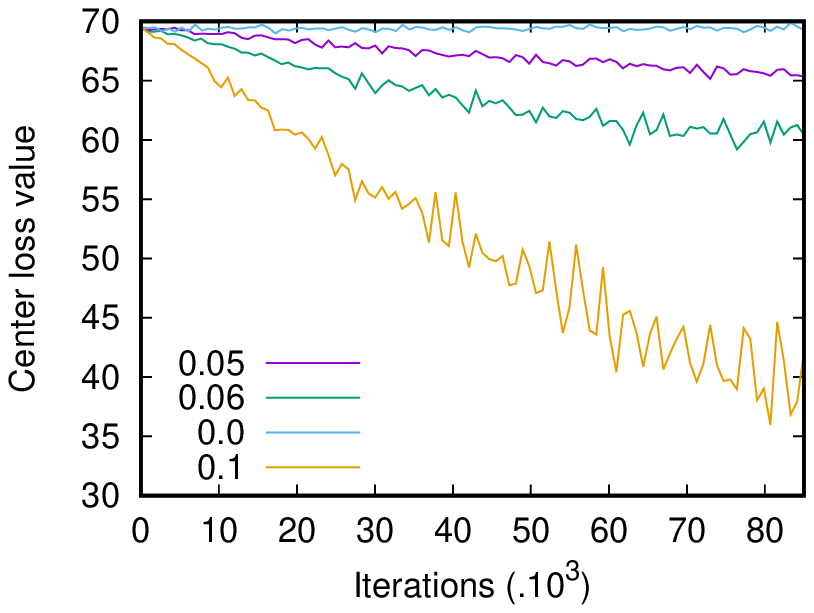}\label{fig:cs_train}}
  \hspace{0.7cm}
   \caption{Center loss values and CMC rank1 values during the training of our model.}\label{fig:train_cs}
\end{figure}

We focus in this section on performance without the attributes loss. The results are the ones in the \emph{No Attributes} columns from the Tables~\ref{fig:ana_cs} . To interpret the results we draw on 
Figures~\ref{fig:cmc_train} (resp. Figure~\ref{fig:cs_train}) the value of the rank-1 CMC during the training (resp. the value of the center loss during the training). The rank-1 CMC during training is computed with all the
datasets combined. We thus compute the CMC against all the identities of all the datasets. We control the regularization of our network to have a rank-1 CMC close to $70$. Higher values lead to a rapid overfitting as we will show with a center loss value ($\alpha$) set to $0.1$.

\begin{table}[ht]
  \centering
 \caption{Comparison of four different center loss values ($\alpha$) with and without the attributes losses. We note a X on the 0.1 row since we are on overfitting regime and the value are therefore low.}\label{fig:ana_cs}
\subfloat[VIPeR]{\label{fig:viper_cs}
\begin{tabular}{ l c c }
\hline
Center loss & No attributes &  attributes \\
\hline
0.0 & 33.1 & 34.3\\
0.05 & 32.5 & 31.7  \\
0.06 & 37.6 & 38.2\\
0.1 & X & 28.6\\
\hline
\end{tabular}
}
  \hspace{0.7cm}
  \subfloat[CUHK01]{\label{fig:cuhk01_cs}

    \begin{tabular}{ l c c }
\hline
Center loss & No attributes & Attributes \\
\hline
0.0 & 68.6 & 67.6\\
0.05 & 64.1 & 63.0  \\
0.06 & 68.7 & 69.7\\
0.1 & X & 42.2\\
\hline
\end{tabular}
}
  \hspace{0.7cm}
  \subfloat[CUHK03]{\label{fig:cuhk03_cs}
    
    \begin{tabular}{ l c c }
\hline
Center loss & No Attributes & Attributes \\
\hline
0.0 & 73.6 & 74.1\\
0.05 & 76.3 & 77.0  \\
0.06 & 77.1 & 77.5\\
0.1 & X & 57.8\\
\hline
\end{tabular}
}
  %\vspace{0.5cm}
\label{tab:infl}
\end{table}

The different values of the center loss lead to different performance. After $0.1$ we observe a drop of performance on all the three datasets. During training, the value of the center loss decreases according to the $\lambda$ 
parameter. This effect can be seen figure \ref{fig:cs_train}. With the value of $0.1$ the center loss decreases dramatically. This has a strong influence to the CMC computed on the training datasets showed on Figure~\ref{fig:cmc_train}:
the rank-1 CMC reaches a value near to 1.0 which leads to overfitting.

This shows that  the center loss has a strong effect during the training, putting the features from a same identify close to a same center increases the capacity of the neural network. With our set of hyperparameters the ideal set of 
value of the center loss is lower than $0.1$.

The center loss has not the same impact on all the datasets. With VIPeR the value increase of  $4.5$ points. For CUHK01 and CUHK03 the value increase of $0.1$ and $4.5$ points. It shows that the center loss really helps the network to output general 
representation of pedestrian. Indeed in the ViPER dataset the variation between the two images of a same class is higher than the ones in the other datasets. Thus the network has to be able of great generalization to perform well on ViPER.

\newpage

\subsection{Influence of the attributes}
We now focus on the attributes. Our objectives are to understand how the attributes help the re-identification score and how our model performs on the attributes extraction task. To compare the influence of the attributes we train
our network with the different values of center loss with the attribute system activated. To activate the attribute we add the factor $100$ on all the attributes losses for our experiments(the $\lambda$ factor). This factor has been empirically found and 
produces the best results. 
As shown in Table \ref{tab:infl} the attributes losses make the network more efficient on all the datasets.

%The curves presented in the following figures represent the CMC on the testing sets - smoothed to have a better vision - trough two trainings : one in blue with the attribute loss activated and the orange one without. One can easily see that by the end of the training, adding attribute in the learning loss makes the network more performant on the Reidentification task. This gain is between one and tow percent depending on the dataset. AJOUTER LES FIGURES DES TRAINS : (cf commentaires suivants)

%"al_0.9_la_0.06_reg_0.001_cla_2919_dp_0.8_ss_4096_bs_64/market-att--inception-PATH14.1-YESdataugNOCropNORatioFine141L2BN099DPATT-2dsMGetand3-image_visu-noMARS-no_art_hums-CPU_eval-Adam-1e-6-attribute-100-299x299" superposé avec :"al_0.9_la_0.06_reg_0.001_cla_2919_dp_0.8_ss_4096_bs_64/market-att--inception-PATH14.1-YESdataugNOCropNORatioFine141L2BN099DPATT-2dsMGetand3-image_visu-noMARS-no_art_hums-CPU_eval-Adam-1e-6--No-attribute-100-299x299"

\subsection{Comparison with the state of the art}

We compare the performance of our system with the state of the art on CUHK01, CUHK03 and VIPeR. Results are shown on Table~\ref{tab:perf}. We take our best model in this Table, i.e our model with the attributes losses enabled and the $\alpha$ value set to $0.06$.

%\newpage

\begin{table}[ht]
  \centering
\caption{CMC top-1 values for different system.}
\begin{tabular}{ l c c c }
\hline
System & CUHK01 & CUHK03 & VIPeR \\
\hline
Best  & 66.6 \cite{guided-dropout} & 75.3 \cite{guided-dropout} & 47.8 \cite{good-viper} \\
Ours & \bf{69.7} & \bf{77.5} & 38.2\\
\hline
\end{tabular}

\label{tab:perf}
\end{table}

Our system shows better performances on CUHK01 and CUHK03 than the state of the art while being  lower on VIPeR. It shows that our approach indeed enables a network to be performant on a large variety of datasets without needing
a special retraining for each of them.

\subsection{Attributes performances}
In this section, the performances of our network will be presented. The tests are run over the test set of the dataset Market-1501. Since some of the classes occur too rarely in the test set to be representative 
(for instance the bottom yellow and purple) they are removed from the tests. The average precision of the network on the attributes recognition are presented in Table \ref{perf_att}. The possible values of the attribute are available on Table~\ref{tab:att}. For both \emph{bottom color} and \emph{top color} attributes, we compute the mean of the average precision of each color. These mean average precision are shown on Table~\ref{perf_att} for the column \emph{bot.col} and \emph{top.col}.

\begin{table}[ht]
 \centering
  \caption{Average precision over the different attributes used in the system}
 \begin{tabular}{| l | c | c | c | c | c | c | c | c | c || c |r| }

 \hline
  \emph{Att}&Gender&len.top &len.bot&len.hair&hand bag& oth.bags&backpack& bot.col&top.col&\emph{mean} \\
  \hline
  \emph{AP}&0.94&0.5&0.97&0.90&0.21&0.54&0.81&0.64&0.80&0.70\\
  \hline
 \end{tabular}

\label{perf_att}
\end{table}

This shows that our system is able to learn to recognize attributes and to perform re-identification at the same time. The attributes classifiers show very good performance on attributes such as gender, hair length, or backpack. Some attributes such as hand bag have a low average precision. This is probably because these classes are under represented on the train dataset. Even if we manage the attribute unbalance during training, when the number of samples for a given class is too limited the network cannot properly generalize.

\section{Conclusions}

In this paper we have shown a CNN architecture for person re-identification. This architecture is trained with multi-task learning in order to have a system able to be trained from different datasets with different labels. We have demonstrated our approach on 4 datasets: two datasets are relatively  small (VIPeR and CUHK01), an another one is larger (CUHK03) and the last one is annotated with attributes (Market 1501). We have evaluated the influence of the different tasks on the global performance (center loss and attributes losses). We proved that combining the different losses leads to better performances. The center loss has a strong influence on performance. We have shown that our system  performs well on CUHK01, CUHK03 and VIPeR, and outperforms recent re-identification works on CUHK01 and CUHK03.

%\section{Acknowledgment}
\clearpage

\bibliographystyle{splncs}
\bibliography{reid}
\end{document}